\documentclass[11pt]{article}
\usepackage[hyperref]{acl2021}
\usepackage[T1]{fontenc}
\usepackage[utf8]{inputenc}
\usepackage{times}
\usepackage{latexsym}
\usepackage{amsmath}
\usepackage{amssymb}
\usepackage{amsfonts}
\usepackage{amsthm}
\usepackage{adjustbox}
\usepackage{booktabs,makecell}
\usepackage{bbm}
\usepackage{bm}
\usepackage{textcomp}
\usepackage{mathtools}
\usepackage{xspace}
\usepackage{graphicx}
\usepackage{makecell}
\usepackage{comment}
\usepackage{scalerel}
\usepackage{marvosym}
\usepackage{bclogo}

\usepackage{empheq}
\usepackage[most]{tcolorbox}

\usepackage{algorithm}
\usepackage[noend]{algpseudocode}
\algnewcommand{\parState}[1]{\State%
    \parbox[t]{\dimexpr\linewidth-\algmargin}{\strut\hangindent=\algorithmicindent \hangafter=1 #1\strut}}

\algrenewcommand\algorithmicindent{1.0em}%

\usepackage{emoji}
\newcommand{\ucambridge}{\text{\emoji[twitter]{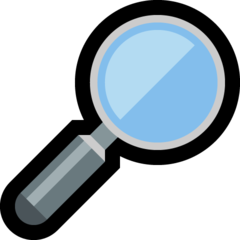}}}
\newcommand{\ethz}{\text{\emoji[twitter]{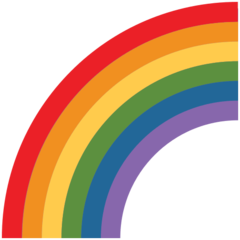}}}

\usepackage{cleveref}
\crefname{section}{\S}{\S\S}
\crefname{table}{Tab.}{}
\crefname{figure}{Fig.}{}
\crefname{algorithm}{Alg.}{}
\crefname{equation}{Eq.}{Eq.}
\crefname{appendix}{App.}{}
\crefname{theorem}{Theorem}{}
\crefname{prop}{Proposition}{}
\crefname{cor}{Corollary}{}
\crefname{observation}{Observation}{}
\crefname{assumption}{Assumption}{}
\crefname{hypothesis}{Hyp.}{Hypotheses}
\crefformat{section}{\S#2#1#3}

\definecolor{darkblue}{RGB}{0,0,160}

\newcommand{\xx}{\mathbf{x}}
\newcommand{\yy}{\mathbf{y}}
\newcommand{\ii}{\mathbf{i}}
\newcommand{\jj}{\mathbf{j}}

\newcommand{\cc}{\mathbf{c}}
\renewcommand{\tt}{\mathbf{t}}
\renewcommand{\ss}{\mathbf{s}}

\newcommand{\LL}{L}
\newcommand{\kernel}{K}
\newcommand{\kernelfunc}{\mathcal{K}}
\newcommand{\calY}{\mathcal{Y}}

\newcommand{\calB}{\mathcal{B}}

\newcommand{\calS}{\mathcal{S}}

\newcommand{\calO}{\mathcal{O}}
\newcommand{\bleu}{\textsc{bleu}\xspace}
\newcommand{\vocab}{\mathcal{V}}
\newcommand{\nmax}{n_{\mathrm{max}}}
\newcommand{\defeq}{\overset{\mathrm{def}}{=}}
\newcommand{\pT}{p_{\scaleto{T}{4pt}}}

\newcommand{\vtheta}{{\boldsymbol \theta}}

\newcommand{\eos}{\textsc{eos}\xspace}
\newcommand{\bos}{\textsc{bos}\xspace}
\newcommand{\defn}[1]{\textbf{#1}}
\newcommand{\tabitem}{~~\llap{\textbullet}~~}
\newcommand{\mathcheck}[1]{{#1}}

\DeclareMathOperator*{\argmax}{argmax}

\newtheorem{prop}{Proposition}

\usepackage[disable]{todonotes}
\makeatletter
\newcommand*\iftodonotes{\if@todonotes@disabled\expandafter\@secondoftwo\else\expandafter\@firstoftwo\fi}  %
\makeatother

\newcommand{\note}[4][]{\todo[author=#2,color=#3,size=\scriptsize,fancyline,caption={},#1]{#4}} %
\newcommand{\ryan}[2][]{\note[#1]{ryan}{violet!40}{#2}}
\newcommand{\clara}[2][]{\note[#1]{clara}{orange}{#2}}

\aclfinalcopy %
\def\aclpaperid{2013}
\title{Determinantal Beam Search}

\author{Clara Meister$^\ethz$~\;~Martina Forster$^\ethz$~\;~Ryan Cotterell$^{\ethz, \ucambridge}$ \\
  $^\ethz$ETH Z\"{u}rich~\;~$^\ucambridge$University of Cambridge \\
   \texttt{meistecl@inf.ethz.ch}~\;~ \texttt{martfors@ethz.ch}\\
   \texttt{ryan.cotterell@inf.ethz.ch}}

\date{}

\begin{document}
\maketitle

\begin{abstract}
    Beam search is a go-to strategy for decoding neural sequence models. The algorithm can naturally be viewed as a subset optimization problem, albeit one where the corresponding set function does not reflect interactions between candidates.
    Empirically, this leads to sets often exhibiting high overlap, e.g., strings may differ by only a single word.  
    Yet in use-cases that call for multiple solutions, a diverse or representative set is often desired. 
    To address this issue, we propose a reformulation of beam search, which we call determinantal beam search. Determinantal beam search has a natural relationship to determinantal point processes (DPPs), models over sets that inherently encode intra-set interactions. By posing iterations in beam search as a series of subdeterminant maximization problems, we can turn the algorithm into a diverse subset selection process. In a case study, we use the string subsequence kernel to explicitly encourage $n$-gram coverage in text generated from a sequence model. We observe that our algorithm offers competitive performance against other diverse set generation strategies in the context of language generation, while providing a more general approach to optimizing for diversity.\looseness=-1

\vspace{1.0em}
    {\includegraphics[width=1.25em,height=1.0em]{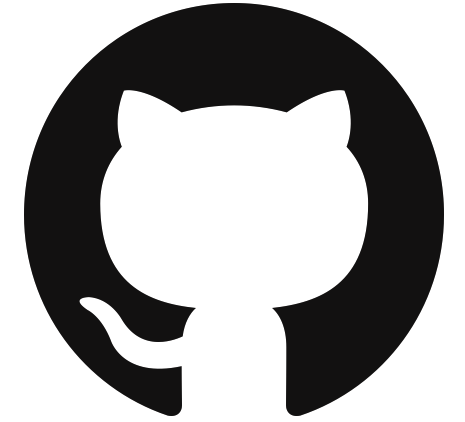}\hspace{1em}\parbox{\dimexpr\linewidth-2\fboxsep-2\fboxrule}{\url{https://github.com/rycolab/swor/tree/diverse_decoding}}}
\vspace{0.1em}
\end{abstract}
\section{Introduction}

The decoding of neural sequence models is a fundamental component of many tasks in NLP. 
Yet, many proposed decoding methods aim to produce only a single solution; further, decoding strategies that provide a set, such as beam search, admit high overlap between solutions.
Such approaches fail to reflect that for many NLP tasks,\footnote{As concrete examples, in machine translation there almost always exist multiple ways to translate a sentence; in story generation, we often seek creative language or multiple options to choose from.\looseness=-1} there can be multiple correct solutions---or that we may desire a diverse set of solutions. 
As it stands, standard beam search chooses items based purely on individual scores, with no means for encoding interaction between candidates; this is the limitation which we attempt to address in this work.

We derive \defn{determinantal beam search}, a novel generalization of beam search that uses the insight that subset selection can be cast as a subdeterminant optimization problem.
Specifically, we formulate each iteration of beam search as a subdeterminant maximization problem parameterized by a positive semi-definite matrix that encodes interactions between the possible candidates; standard beam search is recovered by a specific diagonal matrix. 
This framing creates a natural paradigm for taking the relationships between candidates during the decoding process into account, and can thus assign higher scores to diversified sets; we show how this approach relates to $k$-determinantal point processes (DPPs). 
Given the wealth of research on efficient kernel computation \cite{rousu2005efficient,efficient_kernel} and DPP inference strategies \cite{efficient_kdpp,pmlr-v70-han17a,chen2018fast}, the runtime impact of this method can be quite reasonable in comparison to standard decoding techniques.\looseness=-1

In a case study on neural machine translation (NMT), we demonstrate how to make use of the string subsequence kernel \cite{lodhi2002text} to encode the notion of $n$-gram diversity in the language generation process, allowing us to derive an elegant diverse beam search.
Under this scheme, we observe that determinantal beam search generates more diverse sets than standard beam search with minimal trade-off in terms of \bleu. In terms of this quality--diversity trade-off, we see improved performance over stochastic beam search \cite[SBS;][]{kool2019stochastic}, which is reported to encourage diversity, and a slight improvement over \newcite{vijayakumar2016diverse}'s diverse beam search (DBS) while providing a more general approach to optimizing for intra-set\clara{intra is the correct term here} diversity. 

\section{Neural Sequence Models}\label{sec:prob_gen}
Neural sequence models are probability distributions $p(\yy\!\mid\!\xx)$ over sequences $\yy$ in an output space $\calY$ conditioned on an input $\xx$.\footnote{$\xx$ may be, e.g., a source sentence or an image.} 
Here we define $\calY$ as the set of all valid sequences derived from a vocabulary $\vocab$ that are bookended by distinguished $\bos$ and $\eos$ tokens, indicating the beginning and end of the sequence, respectively. 
Typically, the sequence length is upper-bounded by some value $\nmax\in\mathbb{Z}_+$, which may depend on $\xx$. 
In this work, we consider locally normalized models, i.e. where $p$ is a probability distribution over $\bar\vocab \defeq \vocab \cup \{\eos\}$ conditioned on previously generated tokens $\yy_{<t}$. 
The probability of the full sequence $\yy = \langle y_1, y_2, \dots \rangle$ is then calculated via the chain rule of probability:
\begin{equation}
p(\yy \mid \xx) = \prod_{t=1}^{|\yy|} p(y_t \mid \yy_{<t}, \xx)
\end{equation}
where $\yy_{<1} = y_0 \defeq \bos$. Our model $p$ is typically parameterized by a neural network with weights $\vtheta$. 
As we do not focus on the underlying model itself in this work, we omit the dependence of $p$ on the parameters $\vtheta$. 

We define the \defn{decoding problem} as the search for the highest-scoring $\yy$ among all sequences in $\calY$ according to the model $p(\yy \!\mid \!\xx)$, which is also called maximum-a-posteriori (MAP) inference:
\begin{equation}\label{eq:map}
    \yy^\star = \argmax_{\yy \in \calY}\,\, \log p(\yy \mid \xx)
\end{equation}
\noindent where the $\log$ transform of $p$ is used by convention. 
We further define the \defn{set decoding problem} as the search for a set $Y^\star$ of a specified cardinality $k$ among all valid subsets \mathcheck{$\{Y' \subseteq \calY \mid |Y'| = k\}$} that has the highest score 
where, by overloading, we define 
\begin{equation}
p(Y \mid \xx) \defeq \prod_{\yy \in Y} p(\yy \mid \xx)
\end{equation}
Similarly to \cref{eq:map}, the set-decoding problem is then defined as:
\begin{equation}\label{eq:set_map}
     Y^\star = \argmax_{\substack{Y' \subseteq \calY, \\ |Y'| = k}}\,\, \log p(Y' \mid \xx)
\end{equation}
However, as has been noted in the literature, there are a number of issues with both \cref{eq:map,eq:set_map}. 
First, as $\calY$ may be an exponentially large (in $\vocab$) space and $p$ is typically non-Markovian, we cannot efficiently search over $\calY$, much less over $\calY^k$. 
Second, specifically for language generation tasks, these might not be useful objectives.

\paragraph{Degenerate Objective.}
It is important to note that the highest-probability solutions under neural sequence models are not always high-quality; specifically for tasks involving language generation, e.g., machine translation, prior work has shown the tendency for MAP decoding to lead to generic or degenerate solutions \cite{stahlberg-etal-2019-neural,meister-etal-2020-beam,eikema2020map} while superior solutions assigned only slightly lower probability are often overlooked \cite{Holtzman2020The}. Consequently, heuristic search methods or alternative objectives are frequently employed for decoding language generators.\looseness=-1

\subsection{Beam Search}\label{sec:beam_search}
A common heuristic to approximate the decoding problem in \cref{eq:map} is to sequentially choose the token $y_t$ at each time step $t$ that maximizes $p(y_t \!\mid\! \yy_{< t},\xx)$ until the \eos token is generated or the maximum sequence length $\nmax$ is reached. 
This procedure is known as greedy search.
Beam search is an oft-employed generalization of greedy search that returns $k$ candidates and explores more of the search space.\footnote{A number of NLP tasks only take the highest-scoring element of the returned set $Y$ while other tasks utilize the entire set of solutions.}
In this work, we focus on a framing of beam search
as \defn{iterative subset selection}, which allows for a remarkably
concise formulation of the algorithm.
Given an initial set $Y_0$ containing only the \bos token, we choose subsequent $Y_t$ for $t \in \{1, \dots, \nmax\}$ according to the following recursion:\looseness=-1
\begin{tcolorbox}[ams align,colback=red!5!white,colframe=red!75!black,title=Standard Beam Search]   
Y_0 &\gets \{\bos \} \label{eq:it_subset} \\
    Y_t &\gets \argmax_{\substack{Y'_t \subseteq \calB_t, \\ 
    |Y'_t| = k}}\,\, \log p(Y'_t \mid  Y_{t-1}, \xx) \nonumber
\end{tcolorbox}
\noindent where we are constrained to only extend candidates present in
the \defn{beam set}, which we define as 
\begin{equation}\label{eq:beam_extensions}
    \calB_t \defeq \{ \yy_{<t}\circ y \mid \yy_{<t} \in Y_{t-1} \textbf{ and } y \in \bar\vocab \}
\end{equation}
\noindent where $\circ$ is used to indicate string concatenation.
Note that candidates in $Y_{t-1}$ already ending in \eos are simply added directly to $\calB_t$, i.e., $\eos \circ \eos = \eos$. Under this definition, we have the cardinality constraint \mathcheck{$|\calB_t| \leq |\bar\vocab| \cdot k$}.
\subsection{A Determinantal Reformulation}\label{sec:det}
We now introduce an alternative, equivalent notation for \cref{eq:it_subset} using matrices and determinants that will shed
light on the straightforward generalization of beam search that we
present as the primary contribution of this paper. 
We define a timestep-dependent\footnote{We have omitted the time-step dependence of $D$ for notational brevity as it is always clear from context.}
diagonal matrix \mathcheck{$D\in \mathbb{R}^{|\calB_t| \times |\calB_t|}$} where we take the diagonal entry 
\begin{equation}\label{eq:diagonal-entry}
D_{ii} \defeq p(\yy_{\leq t}^{(i)}\mid \xx)
\end{equation}
\noindent Here $\yy_{\leq t}^{(i)}$ is the $i^{\text{th}}$ candidate in $\calB_t$ according to a unique mapping of every element $\yy_{\leq t} \in \calB_t$ to an integer between 1 and $|\calB_t|$.
Furthermore, we use the notation $D_{Y_t}$ where $Y_t \subseteq \calB_t$, to indicate the submatrix that only contains those rows and columns corresponding to the elements of $Y_t$.
We may now rewrite \cref{eq:it_subset} as 
\begin{tcolorbox}[ams align,colback=red!5!white,colframe=red!75!black,title=Determinantal Standard Beam Search] 
    Y_0 &\gets \{ \bos \} \label{eq:matrix_subset} \\
    Y_t &\gets \argmax_{\substack{Y'_t \subseteq \calB_t, \\ 
    |Y'_t| = k}}\,\, \log \det(D_{Y'_t})\nonumber
\end{tcolorbox}
\noindent where equivalence follows from the definition of the determinant for diagonal matrices. Formally, \cref{eq:matrix_subset} is known as the \defn{subdeterminant maximization problem}\footnote{Albeit with a cardinality constraint.} \cite{og_subdetermmax,subdeterm-max}, which---as the name suggests---refers to the problem of finding the determinant maximizing subset of a matrix.
While the notation introduced in \cref{eq:matrix_subset} may seem contrived, it allows us to perform the subsequent generalization.\looseness=-1

\section{Determinantal Beam Search}\label{sec:determinantal_beam_search}
We are now in a position to ask the fundamental question of this work:
What happens if we replace the diagonal matrix $D$ with a non-diagonal matrix?
This substitution allows us to account for interactions between the elements in the beam.
Formally, we consider a timestep-dependent positive semi-definite (PSD) matrix $D + w \cdot K$ where the off-diagonal matrix $K$ indicates the strength of the interactions between candidates. The non-negative weight $w \geq 0$ controls the importance of these 
interactions during the decoding process.
In this case, the beam search recursion becomes:
\begin{tcolorbox}[ams align,colback=blue!5!white,colframe=blue!75!black,title=Full Determinantal Beam Search]
    Y_0 &\gets \{ \bos \} \label{eq:dpp_matrix_subset} \\
    Y_t &\gets \argmax_{\substack{Y'_t \subseteq \calB_t, \\ 
    |Y'_t| = k}}\,\, \log\det(D_{Y'_t} +  w\cdot K_{Y'_t}) \nonumber
\end{tcolorbox}

Clearly, we recover beam search when $w = 0$; however, we can now select subsets based additionally on candidate interactions.
That is, \cref{eq:dpp_matrix_subset} now has an interpretation as a \defn{diversity objective function} \cite{coreset2} when $K$ is chosen wisely. Due to the presence of the $\log$, \cref{eq:dpp_matrix_subset} is only well defined when the matrix $D_{Y} + w\cdot K_{Y}$ is PSD.\footnote{To see this, recall that the determinant is the product of the eigenvalues. To ensure that the determinant is strictly positive, we can simply enforce that all the eigenvalues are positive, which is necessarily the case for PSD matrices. Note that in the case where any of the eigenvalues of a submatrix are zero, we take $\log \det(\cdot) = -\infty$.}

\subsection{Constructing $K$}\label{sec:cons_k}
One simple way to construct $K$ is as a \defn{Gram matrix}, where each $i,j$ element of $K$ is computed via a kernel function $\kernelfunc: \calS \times \calS \rightarrow \mathbb{R}$ that maps two items in a space $\calS$ to a real number. Specifically, we define $K_{ij} = \kernelfunc(s_i, s_j)$ where
$s_i, s_j \in \calS$ are the $i^{\text{th}}$ and $j^{\text{th}}$ elements of $\calS$, respectively.\ryan{Check this bit!}
In slight abuse of notation, we overload the kernel function $\kernelfunc$ to take a set $S$ such that $K = \kernelfunc(S)$ is the kernel matrix resulting from pairwise computation over elements of $S$.\footnote{In machine learning literature,  the term ``kernel'' is often used to refer to both the function $\kernelfunc$ and the kernel matrix $\kernel$.} 
Following from Mercer's theorem, the matrix $\kernel = \kernelfunc(S)$ is necessarily PSD and, thus the matrix $D_{Y} + w\cdot \kernel_{Y}$ is PSD for any $Y \subseteq S$.\footnote{To see this, note that the matrix $D$ is necessarily PSD. Since PSD matrices are closed under addition and
multiplication by a positive scalar, then necessarily $D + w \cdot K$ is PSD. Lastly, any submatrix of a PSD matrix is also PSD, which makes $D_{Y} + w\cdot \kernel_{Y}$ a PSD matrix.} 

The efficient computation of kernel functions is a well-studied problem---largely due to the prevalence of kernels in various machine learning techniques. For example, dynamic programming techniques are often employed in computation of $\kernelfunc(S)$ \cite{rousu2005efficient} or approximate low-rank kernel matrices can be used in place of $\kernelfunc(S)$ \cite{mem_eff_ker_approx}.\looseness=-1

%
%
%
%
%
%
%
%
%
%
 %
 %
 %
 %
 %
 %
  %
    %
%
%
%
%
%
%
%
%
%
%
%
%
%
%
%
%
%
%
%
%
%
%

\subsection{Relation to a DPPs}
One interpretation of \cref{eq:dpp_matrix_subset} is as a determinantal point process (DPP).
Specifically, it is a $k$-DPP \cite{kulesza2011kdpp} in the $L$-ensemble parameterization where
we have $L = D + w \cdot K$. 
This interpretation as a $k$-DPP gives us a very clear understanding of \emph{why} \cref{eq:matrix_subset} yields a diverse beam search. 
The diagonal entries encode quality, which tells how ``good'' each candidate on 
the beam is, while the off-diagonal entries encode how similar two elements are and, thus, how much they should be repulsed.
For an overview of DPPs we refer the reader to \newcite{taskar_dpp}.

\subsection{Computing Log-Determinants}\label{sec:log-det}

Unfortunately, computing the argmax\footnote{We may also sample from the $k$-DPP modeled by \cref{eq:dpp_matrix_subset} rather than taking the approximate mode; this would only require changing the inference algorithm and can be done in a similarly efficient manner \cite{efficient_kdpp}. We focus on deterministic methods in this work as we aim to find the objective maximizing set.} in \cref{eq:dpp_matrix_subset} is an NP-hard problem \cite{kdpp_nphard}. However, as the subdeterminant maximization problem has many applications, there has been much research on efficient algorithms for approximating log-determinants in the context of, e.g., determinantal point processes \cite{NIPS2012_kdpp,pmlr-v70-han17a}.\footnote{As beam search is already a heuristic approach, such an approximation does not have any theoretical implications for the results of our algorithm. }
One such algorithm uses a first-order approximation of the log-determinant function \cite{pmlr-v70-han17a}. 
The work of \newcite{chen2018fast} uses a greedy, iterative approach; by updating the Cholesky factorization of the matrix kernel incrementally, the algorithm reduces inference time to $\calO(k^2|\calS|)$ to return $k$ candidates from set $\calS$. Pseudocode for the latter approach can be found in \newcite{chen2018fast}; pseudocode for the algorithm in log-space---since probabilistic models are often worked with in log-space for numerical stability---can be found in \cref{app:log_space}.\looseness=-1

\subsection{Runtime Analysis}\label{eq:dbs-runtime}
We consider the runtime of selecting $k$ candidates at any given time step in the recursion of \cref{eq:dpp_matrix_subset}. At each time step, we must first construct the matrix $K$. This computation is highly dependent on the set interactions being modeled; as such, let $\calO(c(k))$ be a runtime bound for $K$'s computation when our search uses a beam size of $k$. Once we have constructed our matrix $D + w \cdot K$, we must next select $k$ items. The set of hypotheses at any time step is at most $k|\bar\vocab|$. While as discussed in \cref{sec:log-det}, finding the size-$k$ subset that exactly maximizes \cref{eq:dpp_matrix_subset} has exponential runtime, we assume approximate methods are employed. Using the method given by \newcite{chen2018fast}, approximate MAP inference takes $k^3|\bar\vocab|$ time to return $k$ items from a set of size $k|\bar\vocab|$. Thus, the runtime at each iteration of determinantal beam search under these conditions would be $\calO(c(k) + k^3 |\bar\vocab|)$. 
Note that standard beam search runs in $\calO(k|\bar\vocab|\log(k|\bar\vocab|))$ time at each iteration. As $k$ is generally small ($\leq 20$) and the impact of $c(k)$ can be made reasonable (\cref{sec:cons_k}), the practical increase in runtime is typically only moderate.\looseness=-1

%

%

%

%
%

%

%
%
%
%
%
%
%
%
%

%

%
%
%
%
%
%
%
%
%
%
%
%
%
%
%

%

\section{Case Study: Diverse Beam Search}
\label{sec:kernel_diverse}

We now consider the task of language generation, where our vocabulary $\bar\vocab$ is a set of words and $\calY$ is the set of all valid strings derived from $\bar\vocab$. When the space of our kernel function $\calS = \calB_t$, one simple way of modeling interactions is through a string subsequence kernel \cite{lodhi2002text}.

\subsection{Computing the String Kernel}\label{sec:computation}
The string subsequence kernel, proposed by \citet{lodhi2002text}, is a function over two strings $\ss$ and $\tt$ computed as:\looseness=-1

\begin{align}
\kernelfunc(\ss,\tt)
&= \sum_{u \in \vocab^n} \sum_{\ii : u=\ss[\ii]} \lambda^{l(\ii)} \sum_{\jj :u=\tt[\jj]} \lambda^{l(\jj)}\label{eq:string_kernel} 
\end{align}
\noindent where $\vocab^n$ is the set of all finite strings of length $n$ over the alphabet $\vocab$; $\ii$ (or $\jj$) denotes a vector of indices $\ii=(i_1, \dots, i_{|u|})$ 
where $1<i_1<i_{|u|}\leq |\ss|$; $l(\ii) \defeq i_{|u|}-i_1+1$ is the length of the substring $u$ in $s$; $\lambda \in (0,1]$ is a decay factor which serves as a penalty for gaps within a compared subsequence.

Direct computation of \cref{eq:string_kernel} is exponential in $|\vocab|$, but efficient dynamic programs can be utilized:
In this work, we employ the trie-based methods of \newcite{rousu2005efficient} to compute \cref{eq:string_kernel}.
Under this scheme, the computation of the kernel between two strings $\ss$ and $\tt$ is $\calO(n\cdot M \cdot\log(\max(|\ss|, |\tt|)))$, where $n$ is the chosen subsequence length (a hyperparameter) and $M$ is the number of words that strings $\ss$ and $\tt$ have in common. 
Note that $|\ss|$, and thus $M$, are bounded by the time step $t$. 
Further, we can reuse many of the computations between subsequent decoding rounds due to the iterative nature of both beam search and the subsequence kernel computations. 
Additionally, since the magnitude of \cref{eq:string_kernel} is influenced by the lengths of $\ss$ and $\tt$, we normalize the kernel as follows:
\begin{equation}
    \kernelfunc_{\mathrm{norm}}(\ss,\tt) = \frac{\kernelfunc(\ss,\tt)}{\sqrt{\kernelfunc(\ss,\ss) \cdot \kernelfunc(\tt,\tt)}}\label{eq:string-norm}
\end{equation}

\subsection{Integration into DetBS}\label{sec:build}
The string subsequence kernel gives us a straightforward method for decoding diverse sets of strings from language generators.
We construct the matrix 
\begin{equation}
    D + w\cdot\kernelfunc(\calB_t)
\end{equation}
using the dynamic program mentioned above to compute $\kernelfunc(\calB_t)$. 
Intuitively, we can expect the argmax---i.e., the size $k$ set corresponding to the objective-maximizing submatrix---of $D + w\cdot \kernelfunc(\calB_t)$ to have higher subsequence diversity as $w$ is increased. 
This is perhaps most easily seen when viewing our problem as a $k$-DPP:
if strings $\yy^{(i)}$ and $\yy^{(j)}$ have high overlap, this will be reflected in the matrix $\kernelfunc(\calB_t)$ at position $i,j$. Higher values of $\kernelfunc(\calB_t)_{i,j}=\kernelfunc(\yy^{(i)}, \yy^{(j)})$ lead to lower probability of both $\yy^{(i)}$ and $\yy^{(j)}$ being in the set drawn according to the $k$-DPP parameterized by $D+w\cdot \kernelfunc(\calB_t)$, which follows from the properties of DPPs outlined in \cref{sec:det}. In short, higher values of $\kernelfunc(\yy^{(i)}, \yy^{(j)})$ decrease the value of $\log\det(D_Y + w\cdot\kernelfunc(\calB_t)_Y)$ for sets $Y$ containing both $\yy^{(i)}$ and $\yy^{(j)}$, which makes $Y$ less likely to be chosen in the recursion of \cref{eq:dpp_matrix_subset}.\looseness=-1

\section{Experiments}
In our experiments, we explore the use of determinantal beam search as a diverse decoding strategy for language generation. 
\begin{figure*}
    \centering
    \includegraphics[width=\textwidth]{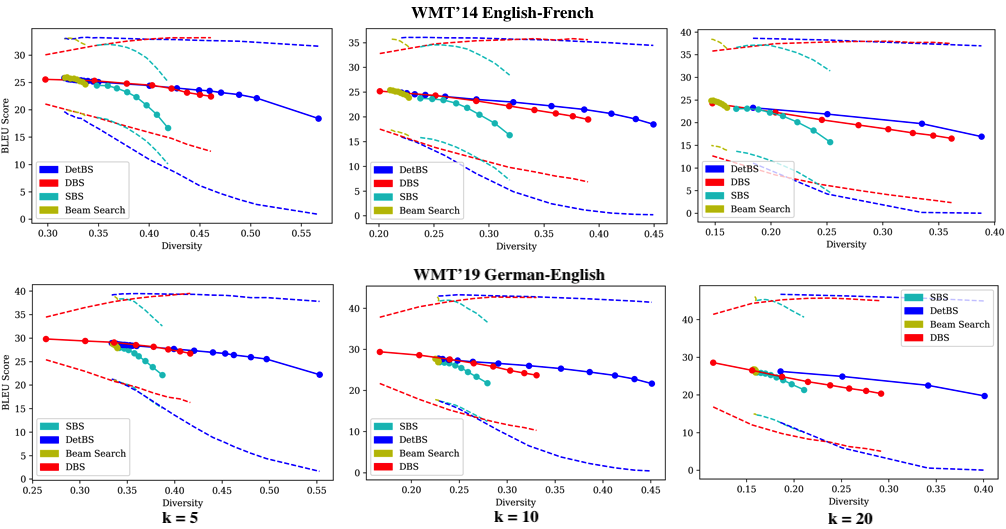}
    \caption{Averaged $n$-gram diversity vs. minimum, median, and maximum \bleu score for beam sizes k = 5, 10, 20 on WMT’14 En--Fr and WMT'19 De--En \texttt{newstest} using various decoding strategies. The free parameter for each strategy is either the softmax temperature or the weight of the diversity parameter (see \cref{sec:setup}). }
    \label{fig:en-fr}
\end{figure*}
\subsection{Baselines}

Various diverse decoding strategies exist in the NLP literature.
We first discuss those strategies that we employ as baselines
in our experiments. 
\paragraph{Standard Beam Search.}
Beam search is one of the most widely used decoding algorithms in NLP, where many problems require efficient strategies for decoding solutions from structured predictors. Specifically, for language generation tasks, beam search has repeatedly proved its effectiveness at decoding state-of-the-art solutions \cite{Wu2016GooglesNM, AAAI1714571, edunov-etal-2018-understanding, XLNET}. We refer back to \cref{sec:beam_search} for the algorithm.

\paragraph{Stochastic Beam Search.}
\citet{kool2019stochastic} propose stochastic beam search (SBS), a decoding technique that samples without replacement from sequence models according to their distribution over the entire space $\calY$.
For random sampling methods such as SBS, it is customary to use a sampling temperature $T > 0$ at generation time to control for the peakiness of the sampling distribution.  
This results in the generalized softmax:
\begin{align}
&\quad\pT(y \mid  \yy_{<t}, \xx) \label{eq:temperature} \\
&\quad\quad = \frac{\exp\Big(\log p(y \mid  \yy_{<t}, \xx)/T\Big)}{\sum_{y' \in \vocab} \exp\Big(\log p(y' \mid  \yy_{<t}, \xx)/T\Big)} \nonumber
\end{align}
\noindent where larger $T$ may lead to more diverse sets simply due to additional smoothing.

\paragraph{Diverse Beam Search.}
\citet{vijayakumar2016diverse} propose a modification to the standard beam search algorithm---which they term diverse beam search (DBS)---to alleviate lack of diversity. The algorithm further divides the beam into $G$ groups $\calB_{t}^1, \dots, \calB_{t}^G$, where $G$ is a hyperparameter of the algorithm, and optimizes for diversity between the different groups by subtracting a similarity term $\Delta(\yy_{\leq t}, \calB_{t}^g)$ from the decoding objective.\footnote{The diversity term has coefficient $w$ to determine the strength of the penalty. When this weight is $0$ or sufficiently small, all groups will return the same solution(s).} Specifically, $\Delta(\yy_{\leq t}, \calB_{t}^g)$ represents the degree of similarity between a hypothesis $\yy_{\leq t}$ and a group of hypotheses $\calB_{t}^g$. 
They find $G=k$, i.e., each group contains a single hypothesis, and the Hamming distance similarity metric lead to the best results; we use these settings in our experiments. Note that under this scheme, the solution set may have duplicates if the diversity penalty is not large enough.\looseness=-1 

Notably, under the above experimental settings, the runtimes of diverse beam search and our algorithm are the same, up to computation of the hamming loss and string kernel, respectively. However, while string kernel computations in our algorithm can be done in parallel, the diversity penalty in diverse beam search must be computed sequentially for each hypothesis, as it is based on the previously chosen groups.

\subsection{Setup}\label{sec:setup}
We run experiments on neural machine translation (NMT) models trained on the WMT'14 \cite{bojar2014findings} En--Fr and the WMT'19 \cite{barrault2019findings} De--En datasets; for reproducibility, we use the pretrained models made available by fairseq\footnote{\url{https://github.com/pytorch/fairseq/tree/master/examples/translation}} \cite{ott2019fairseq}. We evaluate on the \texttt{newstest} set from the respective datasets, each containing 3003 sentences. Further details can be found in \cref{app:exp_setup}.

\begin{table*}
  \centering
  \fontsize{10pt}{12pt}
\selectfont
  \begin{tabular}{l|l }
  \toprule
  \multicolumn{2}{c}{\bf Source Sentence \quad\quad} \\
    Zum Abschluss wurde eine Tombola verlost.&Die Wahrheit zu sagen ist aber kein Verbrechen.\\
      \multicolumn{2}{c}{\bf Beam Search ($T=0.6$) \quad\quad} \\
       \tabitem  \textit{A raffle was held to close the event.}& \tabitem \textit{But telling the truth is not a crime. }\\
      \tabitem  \textit{A raffle was held to conclude the event. }& \tabitem \textit{Telling the truth is not a crime.} \\
      \tabitem \textit{A raffle was held at the end.} & \tabitem \textit{But telling the truth isn't a crime.}  \\
      \tabitem  \textit{At the end a raffle was held.} &\tabitem \textit{But telling the truth is no crime.} \\
      \tabitem \textit{A raffle was held to close the draw.}& \tabitem \textit{However, telling the truth is not a crime.} \\
      
       \multicolumn{2}{c}{\bf Diverse Beam Search ($w=0.4$) \quad\quad} \\
      \tabitem  \textit{To conclude, a raffle was held.} & \tabitem \textit{But telling the truth is not a crime.} \\
      \tabitem  \textit{A raffle was held to close the event.} &\tabitem \textit{But telling the truth is not a crime.} \\
      \tabitem  \textit{A raffle was held to close the event.} & \tabitem \textit{But telling the truth is not a crime.}  \\
      \tabitem  \textit{A raffle was held to close the event.} & \tabitem \textit{But telling the truth is not a crime.} \\
      \tabitem  \textit{At the end of the event, a raffle was held.} & \tabitem \textit{Telling the truth, however, is not a crime.} \\
       \multicolumn{2}{c}{\bf Determinantal Beam Search ($w=0.12$) \quad\quad} \\
      \tabitem  \textit{Finally, a raffle was held.} & \tabitem \textit{But telling the truth is not a crime.} \\
      \tabitem \textit{A raffle was held at the end.} & \tabitem \textit{But telling the truth isn't a crime.}\\
      \tabitem  \textit{At the end a raffle was held.} & \tabitem \textit{Telling the truth is not a crime.}\\
      \tabitem  \textit{To conclude, a raffle was held.} & \tabitem \textit{However, telling the truth is not a crime.} \\
      \tabitem \textit{A raffle was held to close the event.} & \tabitem \textit{But to tell the truth is not a crime.} \\
    \bottomrule
  \end{tabular} 
  \caption{ Generated translations for given source sentence from WMT'19 De--En dataset using different decoding strategies. All use a beam size of five.\protect\footnotemark{}  We see large overlap in Beam Search results while DBS actually returns several of the same results. In comparision, DetBS turns qualitatively diverse results even for simple sentences. }
  \label{tab:examples}
\end{table*}
\footnotetext{For each decoding strategy, we choose the diversity parameter corresponding to the most diverse set that had median \bleu $28.5 \pm 0.05$.}

For determinantal beam search (DetBS), we perform a hyperparameter search (precise details likewise in \cref{app:exp_setup}) over $\lambda$ and $n$, the decay factor and subsequence length, respectively. Search is performed for fixed $w=0.1$ and $k=10$ on validation sets for both languages; we omit a search over the entire space of $w, k, \lambda, n$ so as to not create an unfair advantage for DetBS in comparison with the other decoding strategies, for which no hyperparameters are tuned. We use subsequence length $n=2$ and $\lambda \in \{0.1,0.3\}$ for De--En and En--Fr, respectively.\looseness=-1

We decode sets of size $k \in \{5, 10, 20\}$ with each strategy, comparing sentence-level \bleu and $n$-gram coverage $d_n$ averaged across $n \in \{1,2,3,4\}$ in the decoded sets, where we define $d_n$ as\looseness=-1
\begin{equation}\label{eq:diversity}
    d_n = \frac{\# \textrm{of unique $n$-grams in $k$ strings}}{\# \textrm{of $n$-grams in $k$ strings}}
\end{equation}
While $d_n$ has a more natural interpretation as \emph{coverage} of different $n$-grams, the above quantity is often referred to as $n$-gram diversity in the literature and so we transition to this term for consistency.
Following the experimental setup of \citet{kool2019stochastic}, we vary sampling temperature $T\in \{0.1,0.2,\dots,0.8\}$ in the case of beam search and stochastic beam search and diversity weight $w\in \{0.1,0.2,\dots,0.8\}$ in the case of diverse beam search. For DetBS, we observe that larger sets require a smaller diversity penalty to achieve good $n$-gram diversity: in \cref{fig:en-fr} we show results for DetBS with the string subsequence kernel for $w\in \{0.01, 0.02, \cdots, 0.1, 0.2, 0.3, 0.4\}$ for $k=5$, $w\in \{0.01, 0.02, \cdots, 0.15]$ for $k=10$, and $w\in \{0.01, 0.02, \cdots, 0.05\}$ for $k=20$.\footnote{Recall $w=0$ recovers standard beam search with a temperature of $T=1$.} To observe how \bleu is affected by larger diversity coefficients under DetBS, we explore a finer grain of weights for DetBS in \cref{app:add_resuls}.

\subsection{Results}
\label{sec:nmt_results}

\cref{fig:en-fr} shows the sentence-level \bleu score and averaged $n$-gram diversity on the \texttt{newstest} set for different decoding strategies; \cref{tab:something} shows explicit coverage of $1,2,3,4$-grams and averaged across $1,2,3,4$-grams for different decoding strategies when quality (quantified as \bleu) is controlled for. 
The 3 lines per decoding strategy in \cref{fig:en-fr} represent the minimum, median, and maximum sentence-level \bleu score out of the $k$ translation options, averaged across the corpus. We consider median \bleu to be the best metric of set text-quality, as a good diverse decoding algorithm should not completely sacrifice \bleu for the sake of diversity. The plots are analogous to Fig. 2 in \citet{kool2019stochastic}. 

On both datasets and across different set sizes, results indicate that DetBS generates diverse sets of strings while maintaining high median and maximum \bleu scores. We see similar or higher $n$-gram diversity in comparison to DBS for the same median \bleu and a notably better $n$-gram diversity vs. \bleu trade-off than standard beam search and SBS. Further, the highest quality translation (shown by max \bleu) does not appear to be sacrificed when the diversity parameter is increased for DetBS. In contrast, there is a notable drop-off for generation strategies in which diversity is controlled for using temperature. We show samples of generated text in \cref{tab:examples}.
\begin{table*}
\small
    \centering
    \adjustbox{max width=\textwidth}{
    \begin{tabular}{@{}lllllllllllll@{}} 
        \toprule
         & \multicolumn{6}{c}{\bf De--En} & \multicolumn{6}{c}{\bf En--Fr}  \\
         &\bleu &$1$-gram &$2$-gram &$3$-gram &$4$-gram & avg. &\bleu &$1$-gram &$2$-gram &$3$-gram &$4$-gram &avg. \\
        \hline
        {\bf DetBS} & {\textcolor{darkblue}{\bf 26.24}}&0.11&0.17&0.21&0.26& {\bf 0.19 }& \textcolor{darkblue}{\bf 23.29}&0.11&0.17&0.21&0.25& {\bf 0.18} \rule{0pt}{3ex}\\ 
        {\bf DBS} & \textcolor{darkblue}{\bf 26.52}& 0.09& 0.14&0.18&0.22&0.16 & \textcolor{darkblue}{\bf 24.26}&0.09&0.14&0.17&0.2&0.15\\ 
        {\bf Beam Search}& \textcolor{darkblue}{\bf 26.15}&0.08&0.14&0.19&0.23&0.16 & \textcolor{darkblue}{\bf 23.29}&0.09&0.14&0.18&0.22&0.16 \\ 
        {\bf SBS} &\textcolor{darkblue}{\bf 25.95}&0.08&0.14&0.19&0.24&0.16 & \textcolor{darkblue}{\bf 23.08}&0.1&0.16&0.21&0.25&{\bf 0.18}\\ 
        \bottomrule
    \end{tabular}}
    \caption{We report the coverage (as defined in \cref{eq:diversity}) of $1,2,3,4$-grams and averaged across $1,2,3,4$-grams as well as median \bleu for  $k=20$ on the \texttt{newstest} dataset. For each decoding strategy, we report metrics on the generated set that has highest (average) $d_n$, where we set the constraint that median \bleu for the set is still within 1 point of the highest median \bleu (across decoding strategies and diversity parameters).\protect\footnotemark
    }
    
    \label{tab:something}
\end{table*}

\section{Related Work}
Our work is built upon much of the subset optimization literature in machine learning. We base our algorithm off the subdeterminant maximization problem \cite{coreset}, which has been used to find core sets---a concept originating in computational geometry concerning the existence of a small, representative set of core items---in data summarization problems \cite{data_sum},  nearest neighbor search \cite{nearneighbor} and streaming algorithms \cite{coreset2} \textit{inter alia}.
Informally, we can connect our problem to the notion of decoding a core set from sequence models. To the best of our knowledge, our work is the first to use this concept when decoding sequence models.

\citet{wang} incorporate DPPs into a reinforcement learning objective to optimize for diverse text when training image captioning models. We optimize for diversity during decoding, rather than training, which makes our methods applicable with out-of-the-box models and allows us to avoid highly hyperparameter-sensitive techniques, like minimum-risk training or reinforcement learning-based algorithms, while achieving the same goal. While the application of our methods at training times is an interesting research direction, we foresee technical challenges corresponding to such approaches that may outweigh their benefits.\looseness=-1

There are a number of works that address the problem of diverse decoding in language generation. \citet{weir-etal-2020-cod3s} similarly augment the beam search algorithm, conditioning beams on separate semantic representations to encourage the generation of a diverse set of strings.  \citet{cluster} propose a $k$-means clustering version of beam search that clusters solutions by averaged word embeddings. Perhaps the closest to our work is that of \citet{vijayakumar2016diverse}, who propose a variation of beam search (described in \cref{sec:nmt_results}). However, their algorithm lacks theoretical motivation and is not guaranteed to provide a non-overlapping set;\ryan{This is a good point worth mentioning in the intro} the same solution may appear multiple times in the decoded set if the diversity penalty is not large enough, as shown in \cref{tab:something}. Additionally, groups at each time step $t$ must be processed in order since the score of all hypotheses considered for group $g+1$  depend on hypotheses in groups $1, \dots, g$, which creates a large bottleneck under the recommended settings of $G=k$.\looseness=-1
\footnotetext{In the case that not all strategies had such a set, we instead bounded \bleu by the lowest of the median \bleu across decoding strategies.} 

Random sampling strategies for decoding neural sequence models have received much attention recently.
While techniques such as stochastic beam search and the UniqueRandomizer \cite{shi2020uniquerandomizer} are convenient for creating statistical estimators and have uses in reinforcement learning techniques due to their clear probabilistic interpretation, there are no diversity guarantees for the set of generated sequences.\looseness=-1

\section{Conclusion}

We propose determinantal beam search (DetBS): a new way of framing beam search that allows us to optimize set generation for diversity and coverage rather than simply individual scores. Formally, we redefine beam search as an iterative subdeterminant maximization problem where we select the approximately maximizing set according to the PSD matrix parameterizing our score function. This gives us the ability to encode the notion of intra-set diversity into the beam search optimization problem. 
We discuss and experiment with efficient methods for inference and kernel computation that make DetBS an efficient decoding strategy in practice. 
We use DetBS in the context of language generation, where we explicitly encourage $n$-gram coverage through the string subsequence kernel. In our NMT experiments, we find DetBS generates much more diverse sets of strings than standard beam search and stochastic beam search with a small trade-off in median \bleu. We observe competitive performance compared with diverse beam search.\looseness=-1

\section*{Acknowledgements}
We would like to thank the anonymous reviewers for their helpful feedback and recommendations.

\section*{Ethical Considerations}
While language generation can be used for malicious purposes, e.g., to propagate misinformation or offensive text, we do not foresee any specific ethical concerns with the techniques in this work.\looseness=-1 
\bibliography{acl2020}

\begin{thebibliography}{37}
\expandafter\ifx\csname natexlab\endcsname\relax\def\natexlab#1{#1}\fi

\bibitem[{Abbar et~al.(2013)Abbar, Amer-Yahia, Indyk, Mahabadi, and
  Varadarajan}]{nearneighbor}
Sofiane Abbar, Sihem Amer-Yahia, Piotr Indyk, Sepideh Mahabadi, and Kasturi~R.
  Varadarajan. 2013.
\newblock \href {https://doi.org/10.1145/2462356.2462401} {Diverse near
  neighbor problem}.
\newblock In \emph{Proceedings of the Twenty-Ninth Annual Symposium on
  Computational Geometry}. Association for Computing Machinery.

\bibitem[{Agarwal et~al.(2004)Agarwal, Har-Peled, and Varadarajan}]{coreset}
Pankaj~K. Agarwal, Sariel Har-Peled, and Kasturi~R. Varadarajan. 2004.
\newblock \href {https://doi.org/10.1145/1008731.1008736} {Approximating extent
  measures of points}.
\newblock \emph{Journal of the Association for Computing Machinery},
  51(4):606–635.

\bibitem[{Barrault et~al.(2019)Barrault, Bojar, Costa-juss{\`a}, Federmann,
  Fishel, Graham, Haddow, Huck, Koehn, Malmasi, Monz, M{\"u}ller, Pal, Post,
  and Zampieri}]{barrault2019findings}
Lo{\"\i}c Barrault, Ond{\v{r}}ej Bojar, Marta~R. Costa-juss{\`a}, Christian
  Federmann, Mark Fishel, Yvette Graham, Barry Haddow, Matthias Huck, Philipp
  Koehn, Shervin Malmasi, Christof Monz, Mathias M{\"u}ller, Santanu Pal, Matt
  Post, and Marcos Zampieri. 2019.
\newblock \href {https://doi.org/10.18653/v1/W19-5301} {Findings of the 2019
  conference on machine translation}.
\newblock In \emph{Proceedings of the Fourth Conference on Machine
  Translation}, pages 1--61. Association for Computational Linguistics.

\bibitem[{Bojar et~al.(2014)Bojar, Buck, Federmann, Haddow, Koehn, Leveling,
  Monz, Pecina, Post, Saint-Amand, Soricut, Specia, and
  Tamchyna}]{bojar2014findings}
Ond{\v{r}}ej Bojar, Christian Buck, Christian Federmann, Barry Haddow, Philipp
  Koehn, Johannes Leveling, Christof Monz, Pavel Pecina, Matt Post, Herve
  Saint-Amand, Radu Soricut, Lucia Specia, and Ale{\v{s}} Tamchyna. 2014.
\newblock \href {https://doi.org/10.3115/v1/W14-3302} {Findings of the 2014
  workshop on statistical machine translation}.
\newblock In \emph{Proceedings of the Ninth Workshop on Statistical Machine
  Translation}, pages 12--58. Association for Computational Linguistics.

\bibitem[{Chen et~al.(2018)Chen, Zhang, and Zhou}]{chen2018fast}
Laming Chen, Guoxin Zhang, and Eric Zhou. 2018.
\newblock \href
  {https://papers.nips.cc/paper/2018/hash/dbbf603ff0e99629dda5d75b6f75f966-Abstract.html}
  {Fast greedy map inference for determinantal point process to improve
  recommendation diversity}.
\newblock In \emph{Advances in Neural Information Processing Systems}, pages
  5622--5633.

\bibitem[{Ebrahimi et~al.(2017)Ebrahimi, Straszak, and Vishnoi}]{subdeterm-max}
J.~B. Ebrahimi, D.~Straszak, and N.~K. Vishnoi. 2017.
\newblock \href {https://doi.org/10.1109/FOCS.2017.98} {Subdeterminant
  maximization via nonconvex relaxations and anti-concentration}.
\newblock In \emph{2017 IEEE 58th Annual Symposium on Foundations of Computer
  Science (FOCS)}, pages 1020--1031. IEEE Computer Society.

\bibitem[{Edunov et~al.(2018)Edunov, Ott, Auli, and
  Grangier}]{edunov-etal-2018-understanding}
Sergey Edunov, Myle Ott, Michael Auli, and David Grangier. 2018.
\newblock \href {https://doi.org/10.18653/v1/D18-1045} {Understanding
  back-translation at scale}.
\newblock In \emph{Proceedings of the 2018 Conference on Empirical Methods in
  Natural Language Processing}, pages 489--500. Association for Computational
  Linguistics.

\bibitem[{Eikema and Aziz(2020)}]{eikema2020map}
Bryan Eikema and Wilker Aziz. 2020.
\newblock \href {https://doi.org/10.18653/v1/2020.coling-main.398} {Is {MAP}
  decoding all you need? {T}he inadequacy of the mode in neural machine
  translation}.
\newblock In \emph{Proceedings of the 28th International Conference on
  Computational Linguistics}, pages 4506--4520. International Committee on
  Computational Linguistics.

\bibitem[{Farhan et~al.(2017)Farhan, Tariq, Zaman, Shabbir, and
  Khan}]{efficient_kernel}
Muhammad Farhan, Juvaria Tariq, Arif Zaman, Mudassir Shabbir, and Imdad~Ullah
  Khan. 2017.
\newblock \href
  {https://proceedings.neurips.cc/paper/2017/file/ddcbe25988981920c872c1787382f04d-Paper.pdf}
  {Efficient approximation algorithms for strings kernel based sequence
  classification}.
\newblock In \emph{Advances in Neural Information Processing Systems},
  volume~30, pages 6935--6945. Curran Associates, Inc.

\bibitem[{Gehring et~al.(2017)Gehring, Auli, Grangier, Yarats, and
  Dauphin}]{pmlr-v70-gehring17a}
Jonas Gehring, Michael Auli, David Grangier, Denis Yarats, and Yann~N. Dauphin.
  2017.
\newblock \href {http://proceedings.mlr.press/v70/gehring17a.html}
  {Convolutional sequence to sequence learning}.
\newblock In \emph{Proceedings of the 34th International Conference on Machine
  Learning}, volume~70, pages 1243--1252.

\bibitem[{Gillenwater et~al.(2012)Gillenwater, Kulesza, and
  Taskar}]{NIPS2012_kdpp}
Jennifer Gillenwater, Alex Kulesza, and Ben Taskar. 2012.
\newblock \href
  {https://proceedings.neurips.cc/paper/2012/file/6c8dba7d0df1c4a79dd07646be9a26c8-Paper.pdf}
  {Near-optimal {MAP} inference for determinantal point processes}.
\newblock In \emph{Advances in Neural Information Processing Systems},
  volume~25, pages 2735--2743. Curran Associates, Inc.

\bibitem[{Han et~al.(2017)Han, Kambadur, Park, and Shin}]{pmlr-v70-han17a}
Insu Han, Prabhanjan Kambadur, Kyoungsoo Park, and Jinwoo Shin. 2017.
\newblock \href {http://proceedings.mlr.press/v70/han17a.html} {Faster greedy
  {MAP} inference for determinantal point processes}.
\newblock volume~70 of \emph{Proceedings of Machine Learning Research}, pages
  1384--1393.

\bibitem[{Holtzman et~al.(2020)Holtzman, Buys, Du, Forbes, and
  Choi}]{Holtzman2020The}
Ari Holtzman, Jan Buys, Li~Du, Maxwell Forbes, and Yejin Choi. 2020.
\newblock \href {https://openreview.net/forum?id=rygGQyrFvH} {The curious case
  of neural text degeneration}.
\newblock In \emph{Proceedings of the International Conference on Learning
  Representations}.

\bibitem[{Indyk et~al.(2014)Indyk, Mahabadi, Mahdian, and Mirrokni}]{coreset2}
Piotr Indyk, Sepideh Mahabadi, Mohammad Mahdian, and Vahab~S. Mirrokni. 2014.
\newblock \href {https://doi.org/10.1145/2594538.2594560} {Composable core-sets
  for diversity and coverage maximization}.
\newblock In \emph{Proceedings of the Thirty-Third Association for Computing
  Machinery SIGMOD-SIGACT-SIGART Symposium on Principles of Database Systems},
  page 100–108. Association for Computing Machinery.

\bibitem[{Klee et~al.(1995)Klee, Gritzmann, and Larman}]{og_subdetermmax}
V.~Klee, P.~Gritzmann, and D.~Larman. 1995.
\newblock \href {http://eudml.org/doc/131376} {Largest $j$-simplices in
  $n$-polytopes.}
\newblock \emph{Discrete and Computational Geometry}, 13(3-4):477--516.

\bibitem[{Ko et~al.(1995)Ko, Lee, and Queyranne}]{kdpp_nphard}
Chun-Wa Ko, Jon Lee, and Maurice Queyranne. 1995.
\newblock \href {http://www.jstor.org/stable/171694} {An exact algorithm for
  maximum entropy sampling}.
\newblock \emph{Operations Research}, 43(4):684--691.

\bibitem[{Kool et~al.(2019)Kool, Van~Hoof, and Welling}]{kool2019stochastic}
Wouter Kool, Herke Van~Hoof, and Max Welling. 2019.
\newblock \href {https://arxiv.org/pdf/1903.06059.pdf} {Stochastic beams and
  where to find them: The {G}umbel-top-$k$ trick for sampling sequences without
  replacement}.
\newblock In \emph{Proceedings of the International Conference on Machine
  Learning}, pages 3499--3508.

\bibitem[{Kulesza and Taskar(2011)}]{kulesza2011kdpp}
Alex Kulesza and Ben Taskar. 2011.
\newblock \href {https://icml.cc/Conferences/2011/papers/611_icmlpaper.pdf}
  {$k$-{DPP}s: fixed-size determinantal point processes}.
\newblock In \emph{Proceedings of the 28th International Conference on Machine
  Learning}, pages 1193--1200.

\bibitem[{Kulesza and Taskar(2012)}]{taskar_dpp}
Alex Kulesza and Ben Taskar. 2012.
\newblock \href {https://dl.acm.org/doi/book/10.5555/2481023}
  {\emph{Determinantal Point Processes for Machine Learning}}.
\newblock Now Publishers Inc.

\bibitem[{Li et~al.(2016)Li, Jegelka, and Sra}]{efficient_kdpp}
Chengtao Li, Stefanie Jegelka, and Suvrit Sra. 2016.
\newblock \href {http://proceedings.mlr.press/v51/li16f.html} {Efficient
  sampling for $k$-determinantal point processes}.
\newblock volume~51 of \emph{Proceedings of Machine Learning Research}, pages
  1328--1337.

\bibitem[{Li and Eisner(2009)}]{li_eisner_semiring}
Zhifei Li and Jason Eisner. 2009.
\newblock \href {https://www.aclweb.org/anthology/D09-1005} {First- and
  second-order expectation semirings with applications to minimum-risk training
  on translation forests}.
\newblock In \emph{Proceedings of the 2009 Conference on Empirical Methods in
  Natural Language Processing}, pages 40--51. Association for Computational
  Linguistics.

\bibitem[{Lodhi et~al.(2002)Lodhi, Saunders, Shawe-Taylor, Cristianini, and
  Watkins}]{lodhi2002text}
Huma Lodhi, Craig Saunders, John Shawe-Taylor, Nello Cristianini, and Chris
  Watkins. 2002.
\newblock \href {https://www.jmlr.org/papers/volume2/lodhi02a/lodhi02a.pdf}
  {Text classification using string kernels}.
\newblock \emph{Journal of Machine Learning Research}, 2:419--444.

\bibitem[{Meister et~al.(2020)Meister, Cotterell, and
  Vieira}]{meister-etal-2020-beam}
Clara Meister, Ryan Cotterell, and Tim Vieira. 2020.
\newblock \href {https://www.aclweb.org/anthology/2020.emnlp-main.170} {If beam
  search is the answer, what was the question?}
\newblock In \emph{Proceedings of the 2020 Conference on Empirical Methods in
  Natural Language Processing (EMNLP)}, pages 2173--2185. Association for
  Computational Linguistics.

\bibitem[{Mirzasoleiman et~al.(2013)Mirzasoleiman, Karbasi, Sarkar, and
  Krause}]{data_sum}
Baharan Mirzasoleiman, Amin Karbasi, Rik Sarkar, and Andreas Krause. 2013.
\newblock \href
  {https://proceedings.neurips.cc/paper/2013/file/84d2004bf28a2095230e8e14993d398d-Paper.pdf}
  {Distributed submodular maximization: Identifying representative elements in
  massive data}.
\newblock In \emph{Advances in Neural Information Processing Systems},
  volume~26, pages 2049--2057.

\bibitem[{Ng et~al.(2019)Ng, Yee, Baevski, Ott, Auli, and
  Edunov}]{ng-etal-2019-facebook}
Nathan Ng, Kyra Yee, Alexei Baevski, Myle Ott, Michael Auli, and Sergey Edunov.
  2019.
\newblock \href {https://doi.org/10.18653/v1/W19-5333} {{F}acebook {FAIR}{'}s
  {WMT}19 news translation task submission}.
\newblock In \emph{Proceedings of the Fourth Conference on Machine Translation
  (Volume 2: Shared Task Papers, Day 1)}, pages 314--319. Association for
  Computational Linguistics.

\bibitem[{Ott et~al.(2019)Ott, Edunov, Baevski, Fan, Gross, Ng, Grangier, and
  Auli}]{ott2019fairseq}
Myle Ott, Sergey Edunov, Alexei Baevski, Angela Fan, Sam Gross, Nathan Ng,
  David Grangier, and Michael Auli. 2019.
\newblock \href {https://doi.org/10.18653/v1/N19-4009} {fairseq: A fast,
  extensible toolkit for sequence modeling}.
\newblock In \emph{Proceedings of the 2019 Conference of the North {A}merican
  Chapter of the Association for Computational Linguistics (Demonstrations)},
  pages 48--53. Association for Computational Linguistics.

\bibitem[{Rousu and Shawe-Taylor(2005)}]{rousu2005efficient}
Juho Rousu and John Shawe-Taylor. 2005.
\newblock \href {https://jmlr.org/papers/v6/rousu05a.html} {Efficient
  computation of gapped substring kernels on large alphabets}.
\newblock \emph{Journal of Machine Learning Research}, 6:1323--1344.

\bibitem[{Serban et~al.(2017)Serban, Klinger, Tesauro, Talamadupula, Zhou,
  Bengio, and Courville}]{AAAI1714571}
Iulian~Vlad Serban, Tim Klinger, Gerald Tesauro, Kartik Talamadupula, Bowen
  Zhou, Yoshua Bengio, and Aaron Courville. 2017.
\newblock \href
  {https://www.aaai.org/ocs/index.php/AAAI/AAAI17/paper/download/14571/14217}
  {Multiresolution recurrent neural networks: An application to dialogue
  response generation}.
\newblock In \emph{Proceedings of the Thirty-First AAAI Conference on
  Artificial Intelligence}, page 3288–3294. AAAI Press.

\bibitem[{Shi et~al.(2020)Shi, Bieber, and Sutton}]{shi2020uniquerandomizer}
Kensen Shi, David Bieber, and Charles Sutton. 2020.
\newblock \href {http://proceedings.mlr.press/v119/shi20a.html} {Incremental
  sampling without replacement for sequence models}.
\newblock In \emph{Proceedings of the 37th International Conference on Machine
  Learning}, volume 119 of \emph{Proceedings of Machine Learning Research},
  pages 8785--8795. PMLR.

\bibitem[{Si et~al.(2017)Si, Hsieh, and Dhillon}]{mem_eff_ker_approx}
Si~Si, Cho-Jui Hsieh, and Inderjit~S. Dhillon. 2017.
\newblock \href {http://jmlr.org/papers/v18/15-025.html} {Memory efficient
  kernel approximation}.
\newblock \emph{Journal of Machine Learning Research}, 18(20):1--32.

\bibitem[{Stahlberg and Byrne(2019)}]{stahlberg-etal-2019-neural}
Felix Stahlberg and Bill Byrne. 2019.
\newblock \href {https://doi.org/10.18653/v1/D19-1331} {On {NMT} search errors
  and model errors: Cat got your tongue?}
\newblock In \emph{Proceedings of the 2019 Conference on Empirical Methods in
  Natural Language Processing and the 9th International Joint Conference on
  Natural Language Processing (EMNLP-IJCNLP)}, pages 3356--3362. Association
  for Computational Linguistics.

\bibitem[{Tam(2020)}]{cluster}
Yik-Cheung Tam. 2020.
\newblock \href {https://doi.org/https://doi.org/10.1016/j.csl.2020.101094}
  {Cluster-based beam search for pointer-generator chatbot grounded by
  knowledge}.
\newblock \emph{Computer Speech and Language}, 64:101094.

\bibitem[{Vijayakumar et~al.(2018)Vijayakumar, Cogswell, Selvaraju, Sun, Lee,
  Crandall, and Batra}]{vijayakumar2016diverse}
Ashwin Vijayakumar, Michael Cogswell, Ramprasaath Selvaraju, Qing Sun, Stefan
  Lee, David Crandall, and Dhruv Batra. 2018.
\newblock \href
  {https://www.aaai.org/ocs/index.php/AAAI/AAAI18/paper/view/17329} {Diverse
  beam search for improved description of complex scenes}.
\newblock In \emph{AAAI Conference on Artificial Intelligence}, pages
  7371--7379.

\bibitem[{{Wang} and {Chan}(2019)}]{wang}
Qingzhong {Wang} and Antoni~B. {Chan}. 2019.
\newblock \href {http://arxiv.org/abs/1908.04919} {Towards diverse and accurate
  image captions via reinforcing determinantal point process}.
\newblock \emph{CoRR}, abs/1908.04919.

\bibitem[{Weir et~al.(2020)Weir, Sedoc, and Van~Durme}]{weir-etal-2020-cod3s}
Nathaniel Weir, Jo{\~a}o Sedoc, and Benjamin Van~Durme. 2020.
\newblock \href {https://doi.org/10.18653/v1/2020.emnlp-main.421} {{COD3S}:
  Diverse generation with discrete semantic signatures}.
\newblock In \emph{Proceedings of the 2020 Conference on Empirical Methods in
  Natural Language Processing (EMNLP)}, pages 5199--5211. Association for
  Computational Linguistics.

\bibitem[{Wu et~al.(2016)Wu, Schuster, Chen, Le, Norouzi, Macherey, Krikun,
  Cao, Gao, Macherey, Klingner, Shah, Johnson, Liu, Kaiser, Gouws, Kato, Kudo,
  Kazawa, Stevens, Kurian, Patil, Wang, Young, Smith, Riesa, Rudnick, Vinyals,
  Corrado, Hughes, and Dean}]{Wu2016GooglesNM}
Yonghui Wu, Mike Schuster, Zhifeng Chen, Quoc~V. Le, Mohammad Norouzi, Wolfgang
  Macherey, Maxim Krikun, Yuan Cao, Qin Gao, Klaus Macherey, Jeff Klingner,
  Apurva Shah, Melvin Johnson, Xiaobing Liu, Lukasz Kaiser, Stephan Gouws,
  Yoshikiyo Kato, Taku Kudo, Hideto Kazawa, Keith Stevens, George Kurian,
  Nishant Patil, Wei Wang, Cliff Young, Jason Smith, Jason Riesa, Alex Rudnick,
  Oriol Vinyals, Gregory~S. Corrado, Macduff Hughes, and Jeffrey Dean. 2016.
\newblock \href {http://arxiv.org/abs/1609.08144} {Google's neural machine
  translation system: Bridging the gap between human and machine translation}.

\bibitem[{Yang et~al.(2019)Yang, Dai, Yang, Carbonell, Salakhutdinov, and
  Le}]{XLNET}
Zhilin Yang, Zihang Dai, Yiming Yang, Jaime Carbonell, Rusland Salakhutdinov,
  and Quoc~V. Le. 2019.
\newblock \href
  {https://proceedings.neurips.cc/paper/2019/file/dc6a7e655d7e5840e66733e9ee67cc69-Paper.pdf}
  {{XLNet}: Generalized autoregressive pretraining for language understanding}.
\newblock In \emph{Advances in Neural Information Processing Systems},
  volume~32.

\end{thebibliography}
\bibliographystyle{acl_natbib}

\appendix
\clearpage
\newpage
\section{Log-Space Computations}\label{app:log_space}

\algrenewcommand{\algorithmiccomment}[1]{\hskip1em$\rightarrow$ \footnotesize#1 \normalsize}
\begin{algorithm}
\small
 \textbf{Input: } $\LL$: log of PSD matrix\\
 \hspace*{3em} $k$: desired set size
 \begin{algorithmic}[1]
 \Function{greedy\_map\_inference}{$\,$}
 \State $\cc_i = \boldsymbol[\,\boldsymbol], d_i = \LL_{ii}, \ss_i = \boldsymbol[\,\boldsymbol]$ 
 \State $j=\argmax_{i \in \calS} d_i$ 
 \State $Y_g = \{j\}  $
 \While{$|Y_g| \,\, !\!= \,\, k$}
    \For{$i \in \calS \backslash Y_g $}
    \State $s,\mathrm{\,log\_inner}\,\gets\,\, \textsc{logsumexp}( \cc_j, \cc_i, \ss_i, \ss_j)$
    \State $\textsc{func}\gets\,\,\textsc{log\_add}$ {\bf if }  $s < 0$ {\bf else } $\textsc{log\_minus}$
    \If{$\LL_{ji} > \mathrm{log\_inner}$}
        \State  $s \gets$ 1
        \State $e_i = \textsc{func}(\LL_{ji}, \mathrm{log\_inner}) - 0.5\cdot d_j$ 
    \Else
        \State $s\gets -s$
        \State $e_i = \textsc{func}(\mathrm{log\_inner}, \LL_{ji}) - 0.5
        \cdot d_j$ 
    \EndIf
    \State $\cc_i = [\cc_i \quad e_i], d_i = d_i-2\cdot e_i, \ss_i = [\ss_i \quad s]$ 
    \EndFor
    \State$j = \argmax_{i \in \calS \backslash Y_g} d_i, Y_g = Y_g \cup \{j\} $
    
 \EndWhile
 \State \Return $Y_g$
 \EndFunction
 \Function{logsumexp}{$\cc_1, \cc_2, \ss_1, \ss_2$}
 \State $s,\mathrm{\,log\_inner}\,\gets\,\, 1, -\infty$
 \For{$\langle c_1, c_2, s_1, s_2\rangle \in \langle\cc_1, \cc_2, \ss_1, \ss_2\rangle$}
 \State $s' \gets s_1 \cdot s_2$
  \State $\textsc{func}\gets\,\,\textsc{log\_add}$ {\bf if }  $s == s'$ {\bf else } $\textsc{log\_minus}$
  \If{$\mathrm{log\_inner} > c_1 + c_2$}
  \State $\mathrm{log\_inner} \gets \textsc{func}(\mathrm{log\_inner}, c_1 + c_2)$
  \Else
  \State $s = s'$
  \State $\mathrm{log\_inner} \gets \textsc{func}( c_1 + c_2, \mathrm{log\_inner})$
  \EndIf
 \EndFor
 \Return $s,\mathrm{\,log\_inner}$
 \EndFunction
 \end{algorithmic}
 \caption{Fast Greedy MAP Inference with log-space parameterization \cite{chen2018fast}. We transform computations according to \cite{li_eisner_semiring}.}
 \label{alg:log_map_inf}
\end{algorithm}

\section{Experimental Setup}\label{app:exp_setup}
\paragraph{Hyperparameters.}
As we use the string subsequence kernel of section \cref{sec:kernel_diverse} in DetBS, there are a number of hyperparameters that can be adjusted beyond the diversity weight $w$: the decay factor $\lambda$ indicates the degree to which interior gaps are penalized and subsequence length $n$ indicates the length of the considered substrings $u$. For each language, we perform a search over these two hyperparameters for set size $k=10$ and diversity coefficient $w=0.1$ on validation sets. We use a grid search over $n = [2, 3, 4, 5, 6, 7, 8]$ and $\lambda = [0.1, 0.3, 0.5, 0.7, 1.0]$.
We choose the configuration that yields the highest (average $n$-gram diversity)*\bleu, using this configuration in all subsequent experiments. While there may be better performing hyperparameters under different $k$ and $w$, we omit searching over the entire space to create a fairer comparison with the other decoding strategies. 

\begin{table}[ht!]
    \centering
    \adjustbox{max width=\linewidth}{
    \begin{tabular}{@{}lll@{}} 
        \toprule
        Dataset & $n$ & decay $(\lambda)$  \\
        \hline
        WMT'14 En--Fr & 2 & 0.3   \rule{0pt}{3ex}\\ 
        WMT'19 De--En & 2 & 0.1  \\ 
        \bottomrule
    \end{tabular}}
    \caption{Best performing configurations found in search over string subsequence kernel parameters.\looseness=-1}
    \label{tab:parameter_search_configs}
\end{table}
\noindent Interestingly, larger values of $n$ did not improve performance, and were more computationally expensive; small values of $n$ and decay $\lambda$ appear to offer the best \bleu vs. $n$-gram diversity trade-off.\looseness=-1

\paragraph{Dataset and Model Statistics}
We use a convolutional sequence-to-sequence model trained according to \citet{pmlr-v70-gehring17a} on the WMT'14 En--Fr dataset.\footnote{available at \url{http://statmt.org/wmt14/translation-task.html}} Data preprocessing steps, model hyperparameters and baseline performances can be found in their work. We use the pre-trained model checkpoints made available by fairseq at \url{https://github.com/pytorch/fairseq/tree/master/examples/translation}. We use a Transformer-based model trained according to \citet{ng-etal-2019-facebook} on the WMT'19 De--En dataset.\footnote{available at \url{http://www.statmt.org/wmt19/translation-task.html}} Likewise, data preprocessing steps, model hyperparameters and baseline performances can be found in \citet{ng-etal-2019-facebook}. We similarly use the pretrained model checkpoints made available by fairseq.

\section{Additional Results}\label{app:add_resuls}

\begin{figure}[h]
    \centering
    \includegraphics[width=\linewidth]{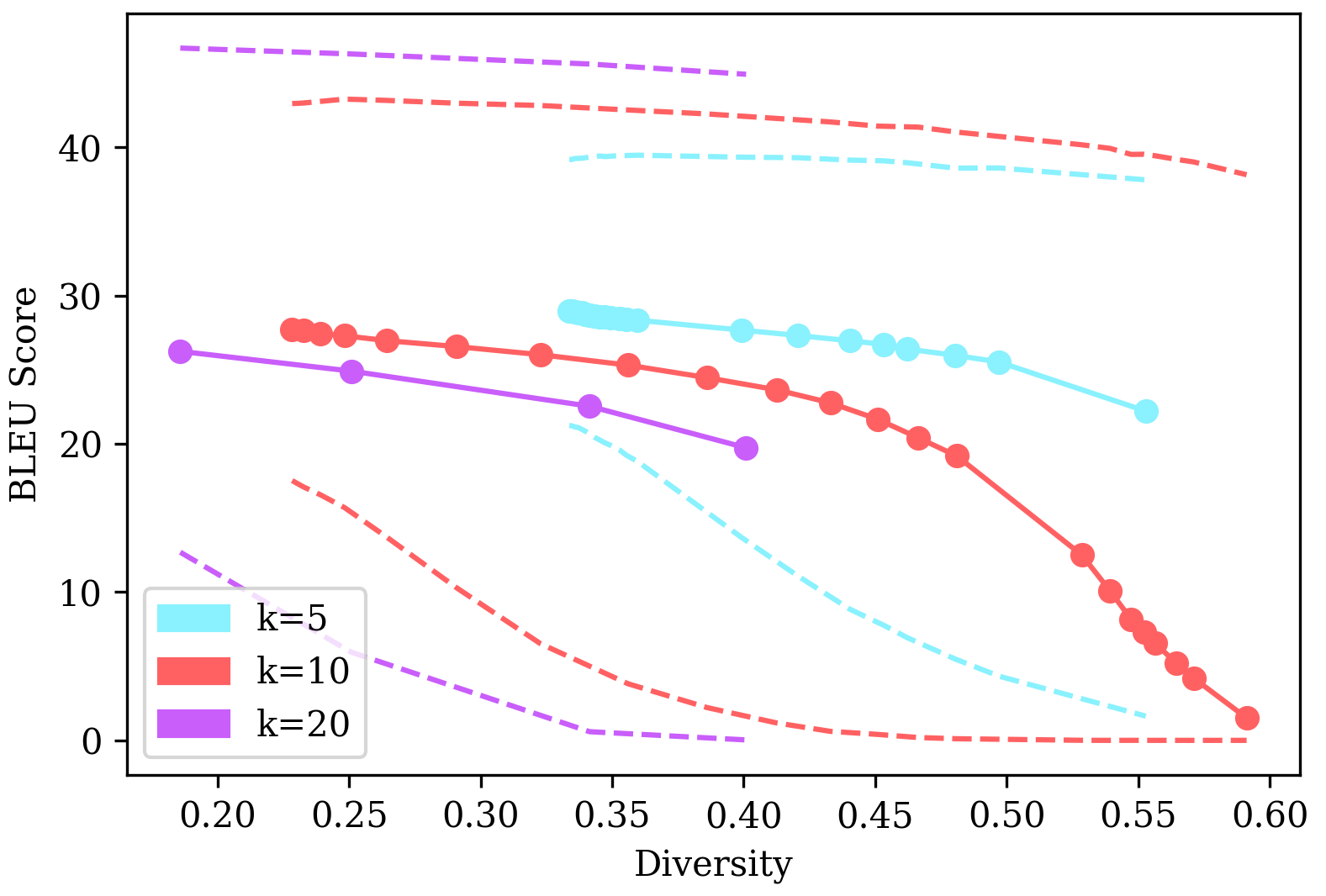}
    \caption{$n$-gram diversity vs. minimum, median and maximum \bleu score for beam sizes k = 5, 10, 20 on WMT'19 De--En \texttt{newstest} using a larger range of the diversity weight $w$.}
    
\end{figure}
\begin{figure}[h]
    \centering
    \includegraphics[width=\linewidth]{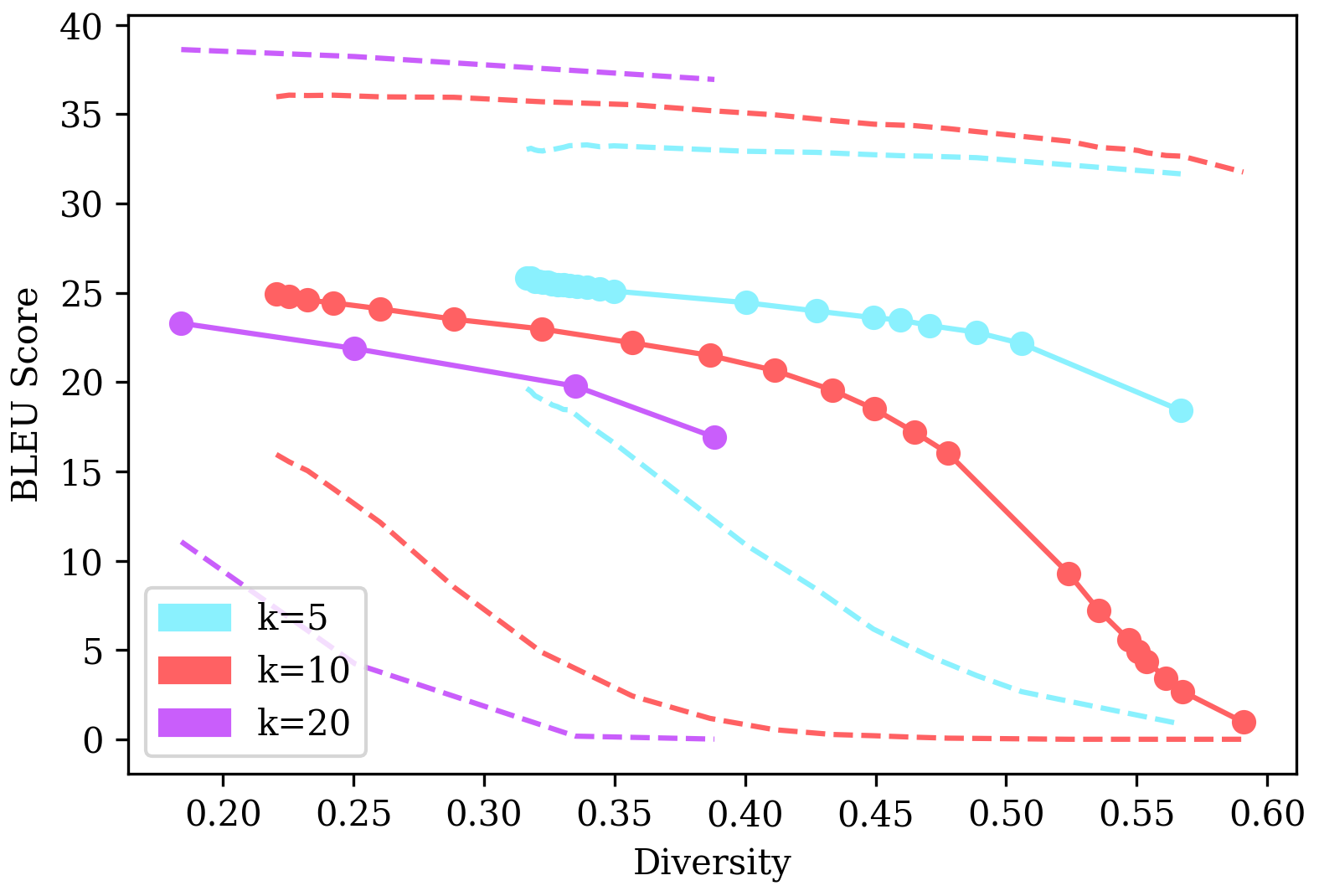}
    \caption{$n$-gram diversity vs. minimum, median and maximum \bleu score for beam sizes k = 5, 10, 20 on WMT’14 En--Fr \texttt{newstest} using a larger range of the diversity weight $w$.}
   
\end{figure}

\end{document}